\documentclass{bmvc2k}
\definecolor{mypink1}{rgb}{0.858, 0.188, 0.478}
\definecolor{ocre}{rgb}{0,0,.4}
\usepackage[font={color=ocre}]{caption}
\usepackage[font={color=ocre}]{subcaption}

\ifdefined\siggraph
\usepackage{times}
\fi

\usepackage{color}
\usepackage{ifthen}
\usepackage{float}
\usepackage{alltt}
\usepackage{newlfont} % for Box
\usepackage{cleveref}
\usepackage{wrapfig}
\usepackage{booktabs}
\usepackage{multirow}
\usepackage{comment}

\usepackage{comment}
\usepackage{amsmath}
\usepackage{amssymb}
\usepackage{amsthm}
\definecolor{SithColor}{rgb}{0.7,0,0} % color for Sith
\definecolor{SetaColor}{rgb}{0.867, 0.0235, 0.376}
\definecolor{ConsularColor}{rgb}{0,0.4,0} % color for the Jedi Consulars (e.g. Yoda)
\definecolor{GuardianColor}{rgb}{0,0,0.8} % color for the Jedi Guardians (e.g. Obiwan)

\definecolor{YajieColor}{rgb}{0.7,0,0}

\definecolor{JunColor}{rgb}{0.5,0,0.7}

\definecolor{WeikaiColor}{rgb}{0.98,0.45,0.0}

\newcommand{\nothing}[1]{}

\definecolor{AudioColor}{rgb}{0.56,0.34,0.62}

\definecolor{DeadlineColor}{rgb}{0.9,0.4,0} % energetic color

\definecolor{figred}{rgb}{1,0,0}
\definecolor{figgreen}{rgb}{0,0.6,0}
\definecolor{figblue}{rgb}{0,0,1}
\definecolor{figpink}{rgb}{1,0.63,0.63}

\newcounter{pccount}
\floatstyle{plain}
\newcommand{\filename}[1]{\url{#1}}
\newcommand{\foldername}[1]{\url{#1}}

\usepackage{graphicx}

\title{Identity Preserving Face Completion for Large Ocular Region Occlusion}

\addauthor{Yajie Zhao}{yajie.zhao@uky.edu}{1}
\addauthor{Weikai Chen}{wechen@ict.usc.edu}{2}
\addauthor{Jun Xing}{junxnui@gmail.com}{2}
\addauthor{Xiaoming Li}{hit.xmshr@gmail.com}{3}
\addauthor{Zach Bessinger}{zach.bessinger@gmail.com}{1}
\addauthor{Fuchang Liu}{20140022@hznu.edu.cn}{4}
\addauthor{Wangmeng Zuo}{cswmzuo@gmail.com}{3}
\addauthor{Ruigang Yang}{ryang@cs.uky.edu}{1}
\addinstitution{
 Computer Science Department\\University of Kentucky\\Lexington, KY, USA
}
\addinstitution{
  Institute for Creative Technologies\\University of Southern California\\Playa Vista, California, USA
}
\addinstitution{
 School of Computer Science and Technology\\Harbin Institute of Technology\\Harbin, China
}

\addinstitution{
 Hangzhou Institute of Service Engineering\\Hangzhou Normal University\\Hangzhou, China
}
\runninghead{Zhao, Yang}{Identity Preserving Face Completion for Large Ocular RO}

\def\eg{\emph{e.g}\bmvaOneDot}

\def\etal{\emph{et al}\bmvaOneDot}

\begin{document}

\maketitle

\begin{abstract}
We present a novel deep learning approach to synthesize complete face images in the presence of large ocular region occlusions. This is motivated by recent surge of VR/AR displays that hinder face-to-face communications. 
Different from the state-of-the-art face inpainting methods that have no control over the synthesized content and can only handle frontal face pose, our approach can faithfully recover the missing content under various head poses while preserving the identity.
At the core of our method is a novel generative network with dedicated constraints to regularize the synthesis process.
To preserve the identity, our network takes an arbitrary occlusion-free image of the target identity to infer the missing content, and its high-level CNN features as an identity prior to regularize the searching space of generator.
Since the input reference image may have a different pose, a pose map and a novel pose discriminator are further adopted to supervise the learning of implicit pose transformations. Our method is capable of generating coherent facial inpainting with consistent identity over videos with large variations of head motions. Experiments on both synthesized and real data demonstrate that our method greatly outperforms the state-of-the-art methods in terms of both synthesis quality and robustness.
\end{abstract}

%-------------------------------------------------------------------------
\section{Introduction}
\label{sec:intro}
Wearable VR/AR devices provide users the ability to travel freely through physical environments mixed with immersive virtual content, enabling new applications in entertainment, education and telepresence. However, the large occlusion introduced by head-mounted display (HMD) is a huge hindrance for face-to-face communications. Such limitation could prevent the adaptation of VR/AR technologies in areas, such as teleconferencing, in which eye contact and facial expressions are crucial elements in effective team communication and negotiation tactics.

\begin{figure}
	\centering
	\newlength{\myheight}
	\newlength{\mywidth}
    \setlength{\myheight}{.13\linewidth}
    \setlength{\mywidth}{.13\linewidth}
\begin{subfigure}[t]{.13\linewidth}
	\includegraphics[width=\mywidth]{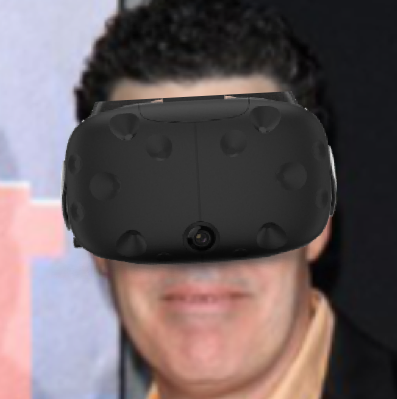}
	\caption{}
	\label{fig:teaser_input1}
\end{subfigure}%
\begin{subfigure}[t]{.13\linewidth}
	\includegraphics[width=\mywidth]{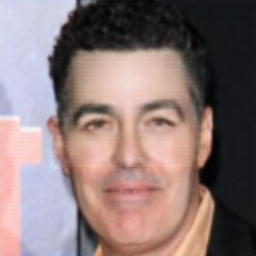}
	\caption{}
	\label{fig:teaser_output1}
\end{subfigure}%
\begin{subfigure}[t]{.13\linewidth}
	\includegraphics[width=\mywidth]{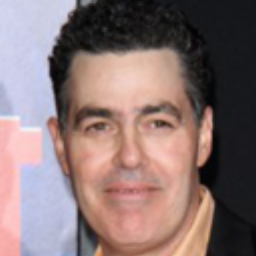}
	\caption{}
	\label{fig:teaser_target1}
\end{subfigure}%
\begin{subfigure}[t]{.11\linewidth}
	\includegraphics[width=\mywidth]{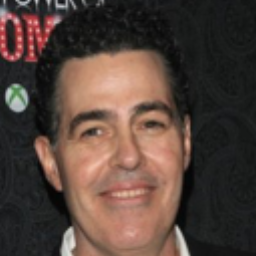}
        \caption{}
        \label{fig:teaser_refs1}
\end{subfigure}
\begin{subfigure}[t]{.11\linewidth}
	\includegraphics[width=\mywidth]{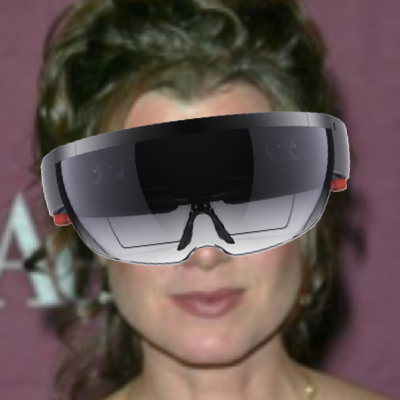}
        \caption{}
        \label{fig:teaser_input2}
\end{subfigure}
\begin{subfigure}[t]{.11\linewidth}
	\includegraphics[width=\mywidth]{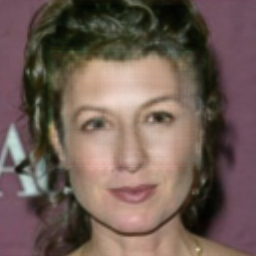}
        \caption{}
        \label{fig:teaser_output2}
\end{subfigure}
\begin{subfigure}[t]{.11\linewidth}
	\includegraphics[width=\mywidth]{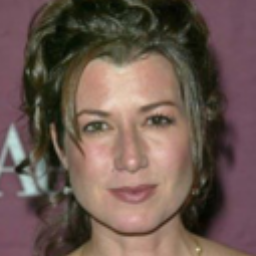}
        \caption{}
        \label{fig:teaser_target2}
\end{subfigure}
\begin{subfigure}[t]{.11\linewidth}
	\includegraphics[width=\mywidth]{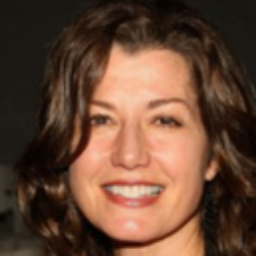}
        \caption{}
        \label{fig:teaser_refs2}
\end{subfigure}
\vspace{0.1in}

\caption{Given a significantly occluded facial image (\subref{fig:teaser_input1}) (\subref{fig:teaser_input2}), we synthesize the un-occluded face image using our framework with identity preserved (\subref{fig:teaser_output1})(\subref{fig:teaser_output2}). (\subref{fig:teaser_target1})(\subref{fig:teaser_target2})show the ground truth image and (\subref{fig:teaser_refs1})(\subref{fig:teaser_refs2}) show the reference images we use of the same person to provide identity information.}
\vspace{-0.08in}
\label{fig:front_page}
\end{figure}

To enable better face-to-face like communications when wearing a HMD, researchers have developed techniques to capture a wearer's expressions to drive a digital avatar~\cite{li2015facial,olszewski2016high,thies2016facevr}. Although some impressive results have been demonstrated, the visual representation is only used for a "talking head" in the VR setting and is limited in quality and details. Instead of driving virtual avatars, another research direction is to inpaint the occluded regions with plausible faces. Nevertheless, inferring the occluded content introduced by VR goggles is particularly challenging as over half of the face is obstructed in the most cases.
~\cite{BurgosArtizzu2015RealtimeEH,zhao2016mask} tried to synthesize the missing texture using personalized database, but requires dedicated capturing setup that makes their methods hard to generalize.

Although some of the state-of-the-art algorithms~\cite{GFC-CVPR-2017} and~\cite{iizuka2017globally} are able to produce plausible face image with large occlusions. Their inputs are required to be frontal, aligned and the identity is usually not preserved during the completion. Such limitations make them infeasible in applications like headset removal, as identity is required to be preserved while the face pose is likely to be changing.

We thus present a novel deep learning approach that can not only fill in the large occluded regions with plausible contents but also provide control over the restored face identity and face poses as shown in Figure~\ref{fig:front_page}. The user could specify the desired face identity by providing an arbitrary occlusion-free face image of the target subject as reference. In addition, by inputting a pose map, our approach could generate facial structures consistent to the intended face orientation.
%The user could specify the desired face pose directly, or control the synthesized identity by providing an occlusion-free face image of the target subject.
These advances are enabled by a generative network that is optimized with  dedicated constraints to regularize the synthesis process.
To inpaint the occluded region with facial content that is visually similar to the input reference image, we introduce a novel reference network that imposes an identity prior onto the searching space of generator. The identity prior is extracted from the referenced identity and penalizes stylistic deviation between the generated result and the input reference image.
At most cases, the reference image is prone to have different pose, illumination and background with the input.
%The control over face pose is achieved through the use of a pose map and a novel pose discriminator.
To obtain a spatially-coherent result, we regularize the generator using two discriminators: a global discriminator that enforces context consistency between filled pixels with surrounding background, and a pose discriminator that regularizes the high-level postural errors.
The pose map serves as both the input of generator and the condition of pose discriminator. By observing the ground-truth face pose, the pose discriminator  penalizes unreal pose transformations produced from the generator.

%Via introducing a novel reference network, our generator is guided to learn the facial identity from the reference image via a perceptual loss \cite{JohnsonAL16} that captures the high-level stylistic feature.
%To get a coherent result, we regularize the synthesis via two adversarial losses: a global loss that enforces context consistency between filled pixels with surrounding background, and a postural loss that penalizes deviated poses from grountruth.

Compared with the previous state-of-the-art methods, our approach is more advantageous in the following aspects. 1) Our method provides significantly better results in the presence of large occluded regions, \eg obstruction from large HMDs. 2) We propose the first face inpainting framework that could explicit control the recovered face identity, which makes identity preserving possible in headset removal. 3) Our approach also offers the editing of face poses in the restored content. To the best of our knowledge, this is the first work that could achieve realistic pose-varying face completion in videos.

\section{Related Work}
\label{sec:relatedwork}

Synthesizing the missing portion of a face image could be formulated as an inpainting problem,
which is first introduced in~\cite{bertalmio2000image}.
To obtain a faithful reconstruction,  content prior is usually required, which comes from either other part of the same image or an external dataset.
The former method generates reasonable inpaintings under specific assumptions, such
as repetitiveness of texture~\cite{efros1999texture}, spatial smoothness in the missing region~\cite{shen2002mathematical} or planar
objects~\cite{huang2014image}. 
However, these methods are prone to fail when completing images with structured content. 
The data-driven methods leverage learnt features from database to infer the missing content ~\cite{hays2007scene,pathak2016context,whyte2009get,mairal2008sparse,xie2012image,yeh2016semantic,fawzi2016image,guptadeeppaint}. 
In particular, the authors in~\cite{hays2007scene,whyte2009get,mairal2008sparse}
generate complete image automatically by using a feature dictionary.

Deep neural network based methods \cite{xie2012image,guptadeeppaint,fawzi2016image} hallucinate the
missing portion of the images by learning through the background texture. 
However, the early attempts tend to generate blurry results and have no control over the semantic meaning of generated result.
More recently, several GAN frameworks have been proposed to address this issue ~\cite{pathak2016context,yeh2016semantic,isola2016image,yang2016high,li2017generative,yeh2017semantic,iizuka2017globally}. 
GANs have been shown to perform well in generating realistic appearing images. ~\cite{li2017generative, iizuka2017globally} solve the general face completion problem by training a model with global and local discriminators. These discriminators ensure that the generated face appear realistic. 
In the face inpainting work of Yeh \etal~\cite{yeh2017semantic}, they search the closet encoding in the latent image manifold to get an inference of how the missing content should be structured, which predicts information in large missing regions and achieve appealing results. 
None of the existing face inpainting approaches is capable of preserving the identity, which makes it infeasible to be applied in the headset removal applications. 
%Only recently has identity-preservation in GANs been addressed~. 

The identity-preserving problem has been explored in related tasks~\cite{li2016deep,huang2017beyond,yin2017towards,tran2017disentangled}, e.g. attributes transfer, frontalization and face recognition.
In particular, pose code has also been introduced in ~\cite{tran2017disentangled, yin2017towards} to resolve the identity ambiguity. 
However, trivially applying the above-mentioned approaches would fail in our case
as none of these works has considered large occlusion, \eg
entirely blocked upper face, in their formulation. 
We show that by jointly learning features from an arbitrary image
of the target identity and a control pose map can significantly
improve the inpainting performance while achieving additional control of face identity and head pose, which, for the first time, enables inpainting over a dynamic sequence
with large head pose variations.

\begin{figure*}[t]
\centering
\includegraphics[width=0.8\linewidth]{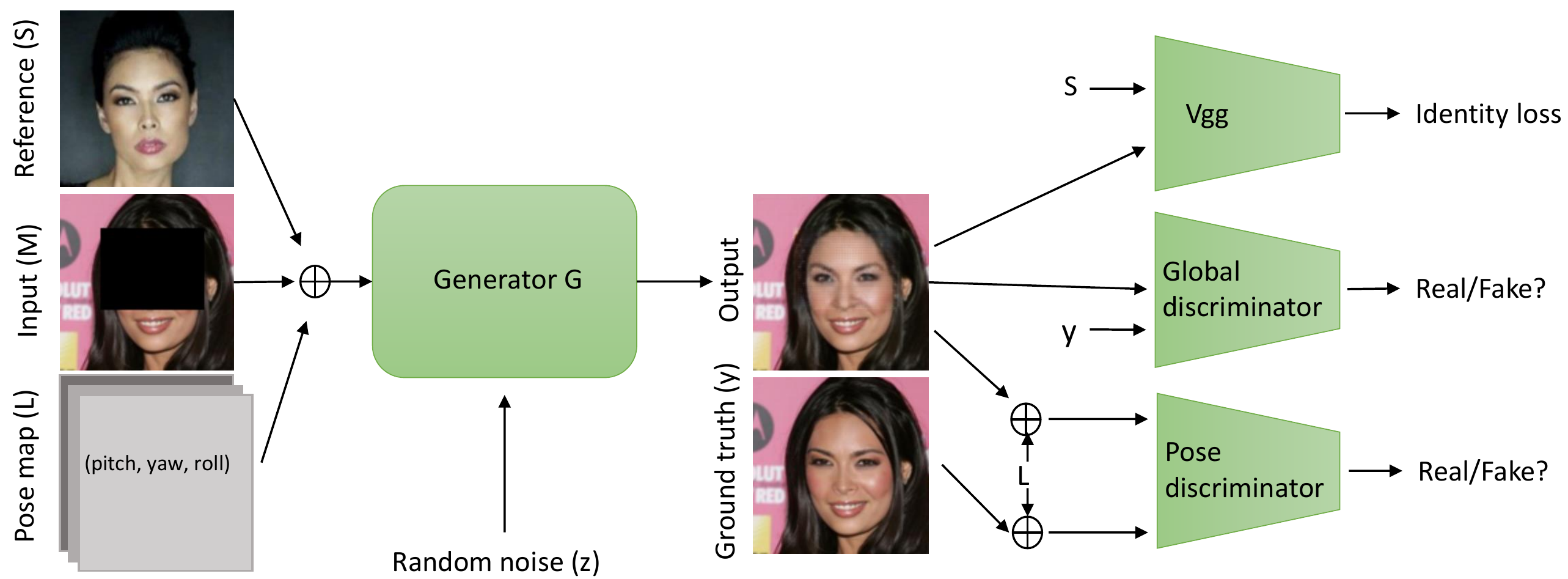}
\vspace{0.1in}
\caption{Our network architecture}
\vspace{-0.1in}
\label{fig:architecture}
\end{figure*}

\vspace{-0.1in}

\section{Identity Preserving Face Completion}
\label{sec:method}

%In this section, we describe the proposed network for identity-preserving face completion. 
To faithfully inpaint the missing region with realistic content while resembling the input identity under a pose constraint, we propose an architecture that consists of a generator and two discriminators as illustrated in Figure~\ref{fig:architecture}. %Figure~\ref{fig:architecture} demonstrates our proposed architecture that consists of a generator and two discriminators.

\vspace{-0.1in}
\subsection{Generator}
\label{sec:generator}

To inpaint large occluded regions with controllable face identity and head pose, the generator $G$ of our network takes three inputs: the occluded face image $\boldsymbol{M}$, an occlusion-free image of a reference identity $\boldsymbol{S}$ and a pose map $\boldsymbol{L}$ that controls the head pose of generated result. The pose map $\boldsymbol{L}$ is a constant-value color image with its three channels encoded by normalized pitch, yaw and roll angles that define the intended face orientation.
Starting from a random variable $\boldsymbol{z}$, we progressively optimize the generator so that it could learn the mapping from a normal distribution to an image manifold $\boldsymbol{\bar{z}}$ that is close to both groundtruth $\boldsymbol{y}$ and the reference image $\boldsymbol{S}$ under the pose constraint $\boldsymbol{L}$.
We formulate the process of finding a recovered encoding $\boldsymbol{\bar{z}}$ as a conditioned optimization problem. In particular, $\boldsymbol{\bar{z}}$ is optimized via solving the following equation:

\begin{equation}
\boldsymbol{\bar{z}} = \mathrm{arg} \min_{\boldsymbol{z}} \{ \mathcal{L}_{r}(\boldsymbol{z} | \boldsymbol{M}, \boldsymbol{S}, \boldsymbol{L}) + \mathcal{L}_{\mathrm{id}}(\boldsymbol{z}, \boldsymbol{S}) \}
\end{equation}
where $\mathcal{L}_{r}$ indicates the reconstruction loss and $\mathcal{L}_{\mathrm{id}}$ denotes an identity loss that penalizes the deviation from the referenced identity.
As $L_2$ loss empirically leads to blurry output and $L_1$ loss performs better on preserving of high-frequency details, we use the $L_1$ loss for measuring the reconstruction error between the generated result and the groundtruth image:

\begin{equation}
\mathcal{L}_{r} = \| \boldsymbol{y} - G(\boldsymbol{z} | \boldsymbol{M}, \boldsymbol{S}, \boldsymbol{L}) \|_1
\end{equation}

Though conditioned on the reference image, only reconstruction loss is not sufficient for ensuring visual similarity with the referenced identity. To achieve identity preservation, we propose to add an identity loss by introducing a \textit{reference network} $R$ that extracts high-level features from the generated result and reference image. We utilize the pre-trained VGG Face network \cite{Parkhi15} as our feature extractor. In particular, we use the $FC6$ feature for both input images. We define the identity loss as the $L_2$ distance between the extracted feature vectors:

\begin{equation}
\mathcal{L}_{\mathrm{id}} = \| \boldsymbol{f}(G(\boldsymbol{z} | \boldsymbol{M}, \boldsymbol{S}, \boldsymbol{L})) -
\boldsymbol{f}(\boldsymbol{S}) \|_2
\end{equation}
where $\boldsymbol{f}$ represents the non-linear feature extracting function learnt by $R$.

\nothing{
To achieve the objective of identity preserving, we introduce an identity sub-net based on the pre-trained VGG Face~\cite{vgg}, which performs well on face recognition. The input to this network is the output of $G$ $I_{fake}=G(M|y,l)$ and an identically-similar reference face, $I_{ref}$. We output the FC6 feature $f\star$ for both input images. The reference network and generator are optimized using the $L_2$ loss as:

\begin{equation}
\mathcal{L}_{\mathrm{ref}} = \sum (f_{I_{\mathrm{fake}}} -
f_{I_{\mathrm{ref}}})^2
\end{equation}
}

% do not be pose aligned; form latent space for projection
%

Note that we do not require the referenced identity image to be pose-aligned with the groundtruth. But our model can still accurately capture the semantic features from the reference image. As demonstrated in Figure~\ref{fig:cross_identity}, the identity-dependent features have been successfully transferred to the generated image. We thus interpret the reference network as a regularizer that imposes an identity/style prior on the manifold of generated images. The proposed network can not only improve the synthesis quality but also stabilize the output to enhance the temporal coherence when dealing with dynamic sequences, e.g. videos.

\nothing{
An important feature of the reference network is that the reference image $\boldsymbol{S}$ is not required to be pose-aligned. Results show that the network is not a copy-paste process from reference image to output as they are not aligned. The use of reference image actually forms an identity latent space during face completion, unlike other face inpainting work only care about the completion of the output, our method will also search for the latent identity space for the closest one. By using of the reference image, we could achieve time spatial consistency when extend of work to video sequences.
}

\nothing{\begin{equation}
\mathcal{L}_{L_1} = \sum | \boldsymbol{x} - G(\boldsymbol{z}|\boldsymbol{y, l}) |
\end{equation}

in which $z$ is the generated completed face. Instead of reconstruction from a noise distribution, $G$ also take the extra condition of reference image and head pose map.}

\nothing{Denote that $x$ as the origin RGB image, $M$ is the occluded version of $x$. $y$ is the reference image of the same identity and $l$ is a spatially same sized grid containing the head pose(pitch, roll and yaw) of $x$. The input to the pose network, $\boldsymbol{y}$,  is a $256 \times 256$
spatial grid of pitch, roll, and yaw extracted using
OpenFace~\cite{amos2016openface}. The grid is laid out such that the first 84
rows are pitch, the second 88 rows are yaw, and the remaining 84 rows are roll. The generator $G$ will take the concatenation of $M$, $y$ and $l$ as input.}

\subsection{Discriminator}
\label{sec:discriminator}

Though the generator can synthesize the missing content with low reconstruction and identity errors, there is no guarantee that the generated image is realistic and consistent with surrounding background. Discriminator serves as a binary classifier that distinguishes real and fake images so that it helps improve the synthesis quality. To encourage photorealism and effective control of face pose, we introduce two discriminators to supervise the generator.

We first introduce a global discriminator $D$ to justify the fidelity and coherence of the entire image. The rationale for introducing a global discriminator is that the inpainted content should not only be realistic but also spatially coherent with surrounding context. In addition, the global discriminator should impose constraints on forming semantic valid facial structures. In particular, we formulate the global discriminator loss function as below:

\nothing{
\begin{equation}
\begin{split}
\mathcal{L}_{\mathrm{global}} = \min _{G} \max _{D}\
&\mathbb{E}_{\boldsymbol{x}
	\sim
	p_\mathrm{data}(\boldsymbol{x})} \big[ \mathrm{log}\
D(\boldsymbol{x} | \boldsymbol{y}) \big] + \\ &\mathbb{E}_{\boldsymbol{z} \sim
	p_{\boldsymbol{z}}(\boldsymbol{z})}\big[ \mathrm{log}\ (1 -
D(G(\boldsymbol{z}|\boldsymbol{y})))\big].
\end{split}
\end{equation}
}
\nothing{
\begin{equation}
\begin{aligned}
\mathcal{L}_{\mathrm{global}} = \min _{G} \max _{D_g}\
&\mathbb{E}_{\boldsymbol{x}
	\sim
	p_\mathrm{data}(\boldsymbol{x})} \big[ \mathrm{log}\
D_g(\boldsymbol{x}, \boldsymbol{M}) \big] + \\ &\mathbb{E}_{\boldsymbol{z} \sim
	p_{\boldsymbol{z}}(\boldsymbol{z})}\big[ \mathrm{log}\ (1 -
D_g(G(\boldsymbol{z}|\boldsymbol{M},\boldsymbol{S},\boldsymbol{L}), \boldsymbol{M}))\big].
\end{aligned}
\end{equation}
}

\begin{equation}
% \resizebox{0.6\hsize}{!}{$
%\begin{split}
\mathcal{L}_{\mathrm{global}} = \min _{G} \max _{D_g}\
\mathbb{E}_{\boldsymbol{x}
	\sim
	p_\mathrm{data}(\boldsymbol{x})} \big[ \mathrm{log}\
D_g(\boldsymbol{x}, \boldsymbol{M}) \big] + \mathbb{E}_{\boldsymbol{z} \sim
	p_{\boldsymbol{z}}(\boldsymbol{z})}\big[ \mathrm{log}\ (1 -
D_g(G(\boldsymbol{z}|\boldsymbol{M},\boldsymbol{S},\boldsymbol{L}), \boldsymbol{x}))\big].
%\end{split}
%$}
\end{equation}

where $p_\mathrm{data}(\boldsymbol{x})$ and $p_{\boldsymbol{z}}(\boldsymbol{z})$ represent the distributions of real data $\boldsymbol{x}$ and noise variables $\boldsymbol{z}$ respectively.
%Both the generator and discriminator are conditioned on the input occluded image $\boldsymbol{M}$ so that the discriminator will penalize the joint configuration of the output, which leads to shaper results.
The global discriminator is sufficient for synthesizing occluded faces with fixed face pose. However, in our application scenario, where the HMD wearer is likely to rotate his/her head while talking, our network should be robust to variable face poses. That means the generated content should have facial structures oriented consistently with the input head pose. We therefore propose an additional pose discriminator $D_{pose}$ to distinguish the faithfulness of synthesized result given the pose constraint. In particular, the pose loss is defined as follows:

\nothing{Unlike the feature aligned database, our database doesn't require semantic parts to be aligned. So we introduced a pose discriminator to distinguish fake/real under the condition of pose map. The loss function is as below:}

\begin{equation}
%\begin{split}
\mathcal{L}_{\mathrm{pose}} = \min _{G} \max _{D_p}\
\mathbb{E}_{\boldsymbol{x}
	\sim
	p_\mathrm{data}(\boldsymbol{x})} \big[ \mathrm{log}\
D_p(\boldsymbol{x}, \boldsymbol{L}) \big] + \mathbb{E}_{\boldsymbol{z} \sim
	p_{\boldsymbol{z}}(\boldsymbol{z})}\big[ \mathrm{log}\ (1 -
D_p(G(\boldsymbol{z}|\boldsymbol{M},\boldsymbol{S},\boldsymbol{L}), \boldsymbol{L}))\big].
%\end{split}
\end{equation}

\nothing{
\begin{equation}
\begin{split}
\mathcal{L}_{\mathrm{pose}} = \min _{G} \max _{D_{pose}}\
&\mathbb{E}_{\boldsymbol{y}
	\sim
	p_\mathrm{data}(\boldsymbol{y})} \big[ \mathrm{log}\
D_{pose}(\boldsymbol{y} , \boldsymbol{L}) \big] + \\ &\mathbb{E}_{\boldsymbol{z} \sim
	p_{\boldsymbol{z}}(\boldsymbol{z})}\big[ \mathrm{log}\ (1 -
D_{pose}(G(\boldsymbol{z}|\boldsymbol{M},\boldsymbol{S} , \boldsymbol{L}),\boldsymbol{L}))\big].
\end{split}
\end{equation}
}

We condition the loss of pose discriminator on the pose map $\boldsymbol{L}$ so that the input pose map would have more accurate control over the inpainted result. However, unlike the global discriminator that back-propagates the gradient over the entire image, the pose discriminator only supervises the loss gradients for the missing region.

\nothing{This pose discriminator uses the pose as a condition to distinguish the real/fake images, which just like discriminators in different categories. The filled part in pixel-wised should have different semantic meaning in different pose categories.}

%\yajie{wrong and redundant}\textcolor{blue}{In addition to the two adversarial losses introduced above, we also propose to add an $L_1$ loss to encourage less blurring.}
Therefore, the overall loss function of our network is defined by:

\nothing{
\begin{equation}
\mathcal{L} = \lambda * \mathcal{L}_{L_1} + \alpha * \mathcal{L}_{\mathrm{global}}
+ \gamma * \mathcal{L}_{\mathrm{pose}},
\end{equation}
}

%\yajie{
\begin{equation}
\mathcal{L} = \lambda * \mathcal{L}_{r} + \mu * \mathcal{L}_{\mathrm{id}} +\alpha * \mathcal{L}_{\mathrm{global}}
+ \gamma * \mathcal{L}_{\mathrm{pose}}
\end{equation}

%}

where $\lambda$, $\mu$, $\alpha$, and $\gamma$ are the weights for the reconstruction loss, identity loss, global discriminator and pose discriminator loss, respectively.

\vspace{-0.1in}
\subsection{Architecture}
\label{sec:archi}

%\yajie{should we make it clear what we actually use}
%Generally speaking, our proposed framework is orthogonal to specific GAN architectures and our approach can take advantage of any generative model.
In our experiment, we adopt U-Net architecture with skipped connections as our generator. Specifically, we concatenate the $i$-th layer onto the $(N - i)$-th layer, where $N$ is the total number of layers, to avoid information loss caused by the bottleneck layer. On the discriminating side, we use PatchGAN for both global and pose discriminator.

\begin{figure*}
%\newlength{\myheight}
%\newlength{\mywidth}
 \setlength{\myheight}{.25\linewidth}
\setlength{\mywidth}{.25\linewidth}
\captionsetup[subfigure]{labelformat=empty}
	\centering
    \setlength{\mywidth}{.11\linewidth}
\begin{subfigure}[t]{.11\linewidth}
	\includegraphics[width=\mywidth]{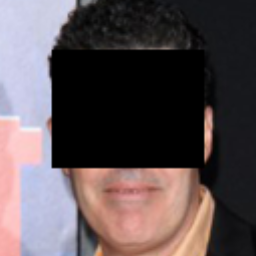}
	\includegraphics[width=\mywidth]{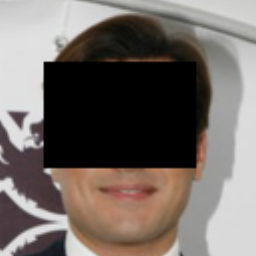}
	\includegraphics[width=\mywidth]{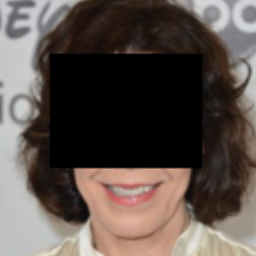}
	\includegraphics[width=\mywidth]{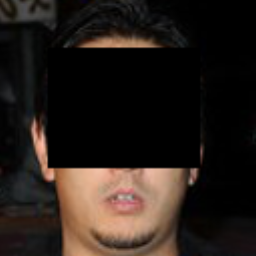}
	\caption{Input}
	\label{fig:teaser_input}
\end{subfigure}%
%\begin{subfigure}[t]{.1\linewidth}
%\begin{subfigure}[t]{.1\linewidth}
\begin{subfigure}[t]{.11\linewidth}
	\includegraphics[width=\mywidth]{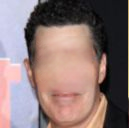}
	\includegraphics[width=\mywidth]{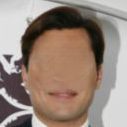}
	\includegraphics[width=\mywidth]{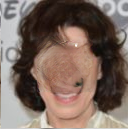}
	\includegraphics[width=\mywidth]{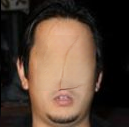}
	\caption{Pathak~\cite{pathak2016context}}
	\label{fig:teaser_target}
\end{subfigure}%
%\begin{subfigure}[t]{.1\linewidth}
\begin{subfigure}[t]{.11\linewidth}
	\includegraphics[width=\mywidth]{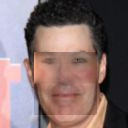}
	\includegraphics[width=\mywidth]{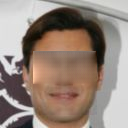}
	\includegraphics[width=\mywidth]{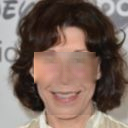}
	\includegraphics[width=\mywidth]{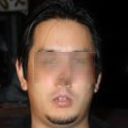}
	 \caption{Yeh~\cite{yeh2017semantic}}
        \label{fig:teaser_ref}
\end{subfigure}
%\begin{subfigure}[t]{.1\linewidth}
\begin{subfigure}[t]{.11\linewidth}
	\includegraphics[width=\mywidth]{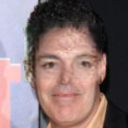}
	\includegraphics[width=\mywidth]{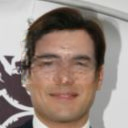}
	\includegraphics[width=\mywidth]{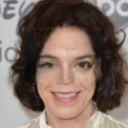}
	\includegraphics[width=\mywidth]{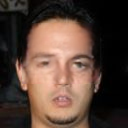}
        \caption{Li~\cite{li2017generative}}
        \label{fig:teaser_ref}
\end{subfigure}
%\begin{subfigure}[t]{.1\linewidth}
\begin{subfigure}[t]{.11\linewidth}
	\includegraphics[width=\mywidth]{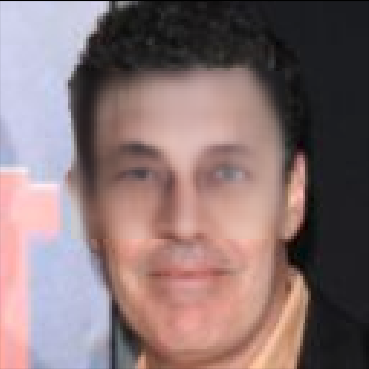}
	\includegraphics[width=\mywidth]{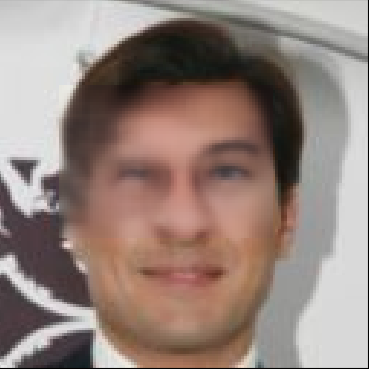}
	\includegraphics[width=\mywidth]{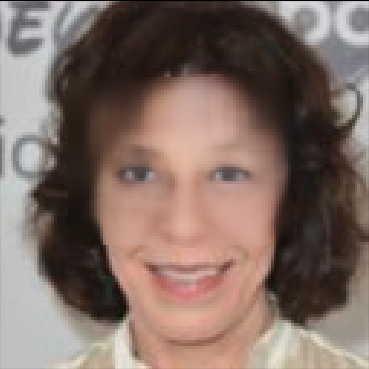}
	\includegraphics[width=\mywidth]{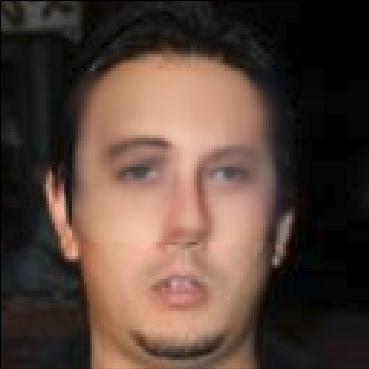}
        \caption{Iizuka~\cite{iizuka2017globally}}
        \label{fig:teaser_ref}
\end{subfigure}
%\begin{subfigure}[t]{.1\linewidth}
\begin{subfigure}[t]{.11\linewidth}
	\includegraphics[width=\mywidth]{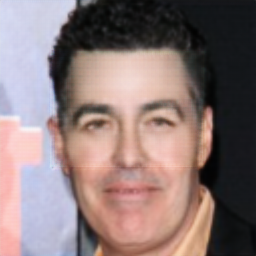}
	\includegraphics[width=\mywidth]{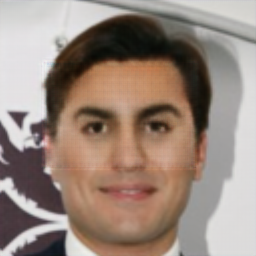}
	\includegraphics[width=\mywidth]{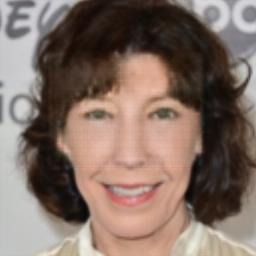}
	\includegraphics[width=\mywidth]{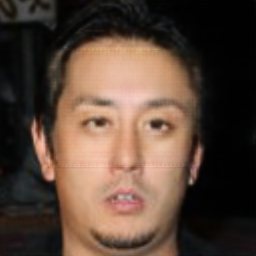}
        \caption{Ours}
        \label{fig:teaser_ref}
\end{subfigure}
%\begin{subfigure}[t]{.1\linewidth}
\begin{subfigure}[t]{.11\linewidth}
	\includegraphics[width=\mywidth]{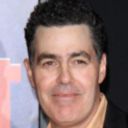}
	\includegraphics[width=\mywidth]{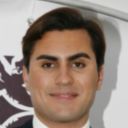}
	\includegraphics[width=\mywidth]{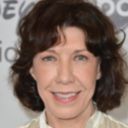}
	\includegraphics[width=\mywidth]{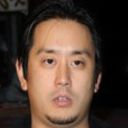}
        \caption{GT}
        \label{fig:teaser_ref}
\end{subfigure}
\vspace{0.1in}
\caption{Comparison with state-of-the-art inpainting frameworks.}
\vspace{-0.1in}
\label{fig:comparison}
\end{figure*}

\vspace{-0.2in}
\section{Implementation Details}
\label{sec:implementation}

We use images from  MS-celeb-1M~\cite{guo2016ms} to construct our training data. Our model is trained using 476 identities and 8000 pair of images (the occluded face image and its reference identity image). To prepare the data, MTCNN~\cite{7553523} is applied to detect the landmarks and bounding boxes. We then scale all the images of the dataset to $128*128$ and align them by registering the nose tip.
The loss functions are optimized using the Adam optimizer, with
a learning rate of 0.0002 and $\beta_1 = 0.5$. We train the network for 100
epochs. Our framework is implemented using Torch~\cite{collobert2011torch7}. In
all experiments, we set our loss hyperparameters as $\lambda = 1$, $\mu =100$,$\alpha =
100$, and $\gamma = 70$.  The momentum is set to 0.9 in our training process. 

\vspace{-0.1in}
\paragraph{Runtime.} We implement our model with Torch~\cite{collobert2011torch7} on a platform of Intel E3 CPU, 3.30GHz and Nvidia GTX-1080 GPU. We can reconstruct face images with size $128\times128$ at a frame rate of 20 Fps. 
%-------------------------------------------------------------------------
\vspace{-0.2in}

\section{Experimental Results}

\label{sec:eval}
\vspace{-0.1in}
\subsection{Face Identity Control}
\label{sec:exp_faceid}

\begin{figure}[t]
	\centering
	\includegraphics[width=0.85\linewidth]{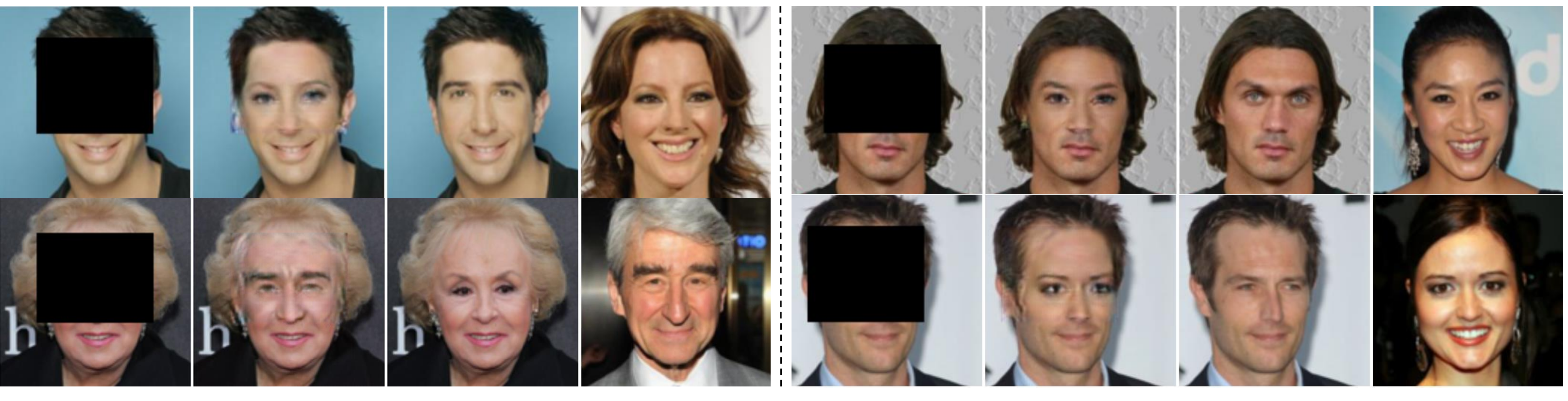}
	\vspace{0.1in}
	\caption{Cross identity experiments. From left to right: inputs, generated image, ground truth, reference images of another identity.}
	\label{fig:cross_identity}
\end{figure}

In this section, we evaluate the effectiveness of the proposed face identity control. Figure~\ref{fig:cross_identity} demonstrates the cross identity experiments, where the image of another identity is fed into the network as reference image. As seen in Figure~\ref{fig:cross_identity}, the referenced identity differs significantly from the original identity in terms of appearance and even genders. 
%different appearance or even different gender with the original identity. 
However, our model can still generate high-fidelity result with spatial coherence while capturing the high-level identity-dependent features, \eg thick eyebrows, eye color, of the referenced identity.

%\textcolor{mypink1}{identical reference for different inputs, probably in suppl.}
\nothing{We further verify the robustness and consistency of our identity control framework. As shown in Figure~\ref{fig:consistense}, we apply the identical face image as the referenced identity for different inputs. The results demonstrate that our network consistently replicates the input target identity regardless of a large variations between the original un-occluded face identities. In addition, the newly inpainted content naturally blends with the original background.}

%In Figure~\ref{fig:cross_identity}, we show interesting results on face completion using reference images from a different identity. The results clearly show the identity attributes impact on the generated images. This identity control power makes our network capable to deal with scenario needs identity coherence like video sequences.

%Our network is robust and consistent as shown in Figure~\ref{fig:consistense}, with the same reference image, even the inputs are different, the outputs are consistent in identity.

\paragraph{Quantitative Analysis} To further quantify the performance of identity preservation, for each synthesized result, we apply the OpenFace~\cite{amos2016openface} to verify the identity similarity of our result compared to the ground-truth and reference image used in our network. In particular, the OpenFace will generate a binary output (0 or 1) for indicating if its two input images capture the same identity. 
We compare the performance of Li \etal~\cite{li2017generative}, Iizuka \etal~\cite{iizuka2017globally}  and our algorithm using 588 images pairs(the input and reference image are the same person). As demonstrated in Table\ref{table:facerecognition}, our method significantly outperforms Li \etal~\cite{li2017generative} and achieves a $93\%$ pass rate when comparing with the groundtruth, indicating the efficacy of our method in preserving the target identity. 
%Note that the pass rate for reference image is slightly lower than that of groundtruth. 
%because the fact that face shape, skin color also play critical role in face verification.
%This is due to the fact that for each synthesized image, the regions outside the mask remain the same with the groundtruth while being different comparing with the reference image.

%conduct face verification using  OpenFace~\cite{amos2016openface} with 588 test pairs(input and reference are the same person). This verification predicts similarity scores of two faces images by computing the distance between their feature vector in face recognition space. For each of the reconstructed image, we compare it with both ground truth and the reference image. Table~\ref{table:facerecognition} shows the verification rates of proposed network and Li \etal~\cite{li2017generative} . The verification rates of our methods significantly outperforms Li \etal . Note that the the verification rate with ground truth is better than with reference for both of the networks. The reason is that the region outside mask contributes to the verification which is unchanged when compared with ground truth.

\begin{table}[]
\centering

\caption{Comparison of Face Verification}
\vspace{0.1in}
\label{table:facerecognition}
\scalebox{0.75}{
\begin{tabular}{cccc}

\hline
&\textbf{ ours (\%) } &\textbf{ Li (\%)  } &\textbf{ Iizuka (\%)  } \\   
\hline
\textbf{Compare with groundtruth} & 93.2 & 57.3 & 45.9  \\
\textbf{Compare with reference} & 87 & 42.8 & 34.7 \\

\hline

\end{tabular}
}

%\end{center}
%]
\vspace{-0.1in}
\end{table}
\begin{figure*}[t]
	\centering
   \includegraphics[width=0.95\linewidth]{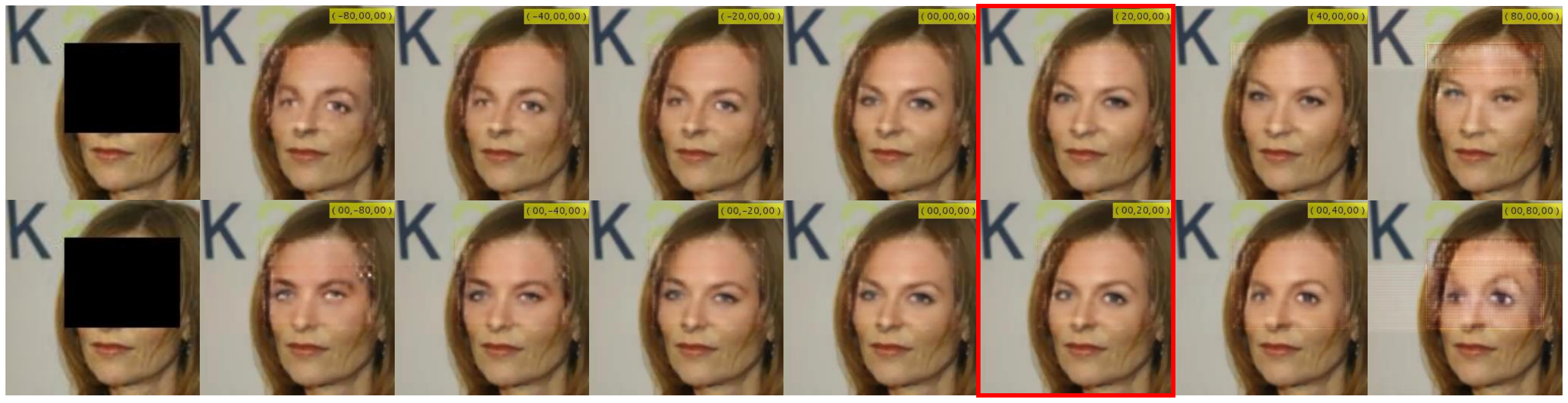}
   \vspace{0.07in}
   \caption{Impacts of pose variation on the face reconstruction.  The ground truth pose of the input image (first column) is $(pitch=15, yaw=20, roll=0)$, and first/second rows show the effects of different pitch/yaw values on the reconstruction results. As we can see, the close poses (in red box) tend to generate visually better results.
   }
\label{fig:Pose}
\end{figure*} 

\begin{figure*}[t]
	\centering
   \includegraphics[width=0.8\linewidth]{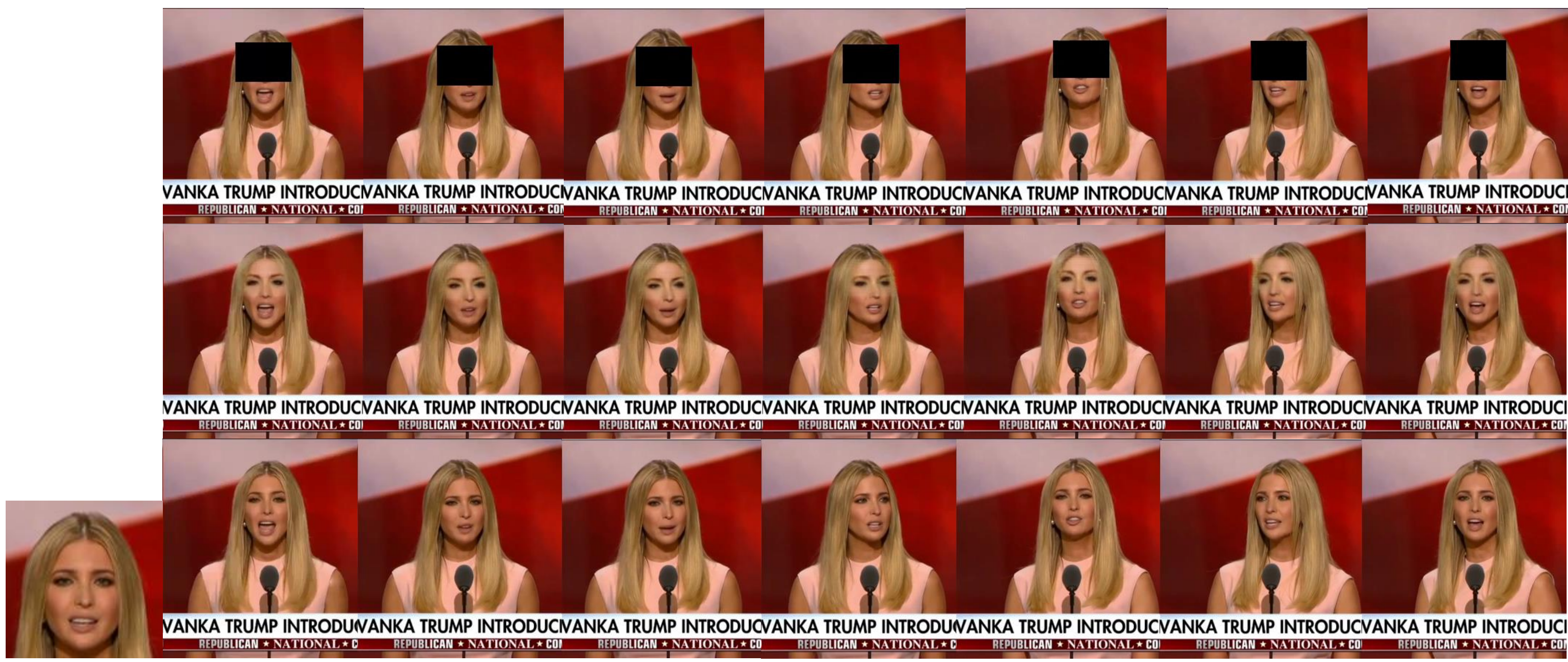}
    \vspace{0.07in}
   \caption{Video example. From top to bottom: input video frames, generated outputs from our network, and the ground truth. The reference image is provided left-most.}
\vspace{-0.2in}
\label{fig:PoseVideo}
\end{figure*}

\vspace{-0.1in}
\subsection{Head Pose Editing}
In this section, we validate the effectiveness of the proposed pose editing component.
Figure~\ref{fig:Pose} shows how the pose variation influences the reconstruction of the face images. By providing different pose inputs, our reconstructed facial structures will be aligned accordingly. The first row of Figure~\ref{fig:Pose} shows the impact of pitch variance on reconstructed results. By gradually increasing the pitch value, the synthesized eye will move up accordingly. The similar control effect is manifested in tuning the yaw values as shown in the second row of Figure~\ref{fig:Pose}. As the face regions outside the mask remain fixed, it is more natural that only the correct pose input will lead to the most coherent and clear results. As seen from our result, only the closer poses (highlighted in red) generates visually better results, suggesting the accuracy of our pose control.

%By providing different pose information, the reconstructed feature will located differently. The head pose is shown on the top right of each image with the format of $(pitch, yaw, roll)$. The first row shows the impact of pitch variance. The horizontal red line indicates the eye position changes just as turning head up and down. The second row shows the changes in yaw as if the head is turning from left to right. We could tell from the region above the translucent mask.

The introduction of pose editing component ensures our model to perform face inpainting in dynamic sequence with time-space coherence. In Figure~\ref{fig:PoseVideo}, we show the reconstructed frames with different head poses from a video sequence by using only one frontal reference. Regardless of the large variations of poses, our network can stably reconstruct appealing results. To better evaluate our method, we provide video results at \href{https://youtu.be/4qGaARE8ob4}{https://youtu.be/4qGaARE8ob4}.

\nothing{
In this section, we show the effectiveness of our pose component.
Figure~\ref{fig:Pose} shows how the pose variation influence the reconstruction of the face images.
The head pose is shown on the top right of each image with the format of $(pitch, yaw, roll)$.
The first row shows the impact of pitch variance. As the pitch value increases, the eye look-at direction changes from up to down. The second row shows the changes in yaw as if the head is turning from left to right.
As we can see, when the pose gets close to the ground truth $(15, 20, 0)$, the quality gets better.

The introduction of pose component makes that possible to do the video editing with time space consistence. In Figure~\ref{fig:PoseVideo}, we show selected reconstructed frames with different head poses from a video sequence by using only one frontal reference. Even with the large poses, our network can still reconstruct appealing results. For video results, please refer to the supplementary material for more details. \nothing{\textit{(Videos can be found in supplementary material)}.}
}

\vspace{-0.1in}
\subsection{Ablative Analysis}

\begin{figure}[ht]
	\centering
   \includegraphics[width=0.95\linewidth]{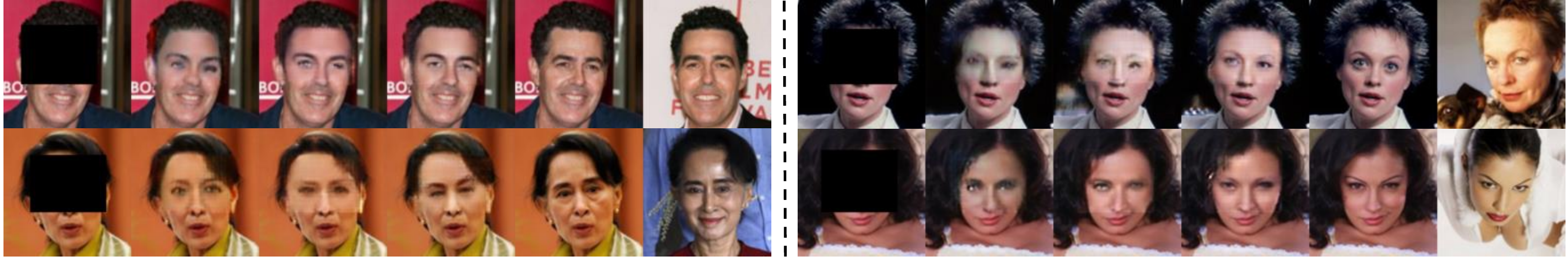}
    \vspace{0.07in}
   \caption{The ablative analysis. From left to right: input, $L_1$+GAN, $L_1$+GAN+ID, $L_1$+GAN+ID+pose, ground truth, and reference image.}
   %\vspace{-0.2in}
\label{fig:allative analysis}
\end{figure}

To access the efficacy of each introduced loss, we experiment on three combinations of losses: $L_1$+GAN, $L_1$+GAN+ID and $L_1$+GAN+ID+Pose.
%, in which is the baseline network inspired by~\cite{isola2016image}.
 %, which is after the introduction of identity loss. , which is the proposed network also including the pose discriminator. 
 Figure~\ref{fig:allative analysis} shows the comparison of the above networks. In general, $L_1$+GAN tends to generate blurry results and fails to capture spatial coherency.% because we use unaligned database.
 $L_1$+GAN+ID demonstrates better performance on preserving the identity, although the result is still blurry and mis-aligned in pose. $L_1$+GAN+ID+Pose, which is our proposed method, is capable to generate images with sharper details while faithfully capturing the referenced identity.
 %and are rich in detail with natural poses.
 The results indicate that the pose control component acts as an implicit alignment prior to register different features to reduce the blurness for each semantic part.
Table~\ref{table:psnr and SSIM} shows the quantitative evaluation on the test set. Our proposed network outperforms the other methods in both PSNR and SSIM.

%\begin{table}[]
%\centering
%
%\caption{\textcolor{mypink1}{Quantitative Evaluation on Each Ablative Networks}}
%\vspace{0.1in}
%\label{table:psnr and SSIM}
%\scalebox{0.75}{
%\begin{tabular}{ccc}
%\hline
%&\textbf{ PSNR } &\textbf{ SSIM  }  \\
%\hline
%\textbf{Li~\cite{li2017generative}} & 21.69 & 0.78 \\
%\textbf{Yeh~\cite{yeh2017semantic}} & 18.87 & 0.79 \\
%\textbf{$L_1$+ GAN + ID} & 23.81 & 0.83 \\
%\textbf{$L_1$+ GAN + ID + Pose} & 24.9 & 0.87  \\
%\hline
%\end{tabular}
%}
%\vspace{-0.1in}
%\end{table}

\begin{table}[]
\centering

\caption{Quantitative Evaluation on different Networks}
\vspace{0.1in}
\label{table:psnr and SSIM}
\scalebox{0.75}{
\begin{tabular}{cccccc}
\hline
&\textbf{ \textbf{Yeh~\cite{yeh2017semantic}} }&\textbf{\textbf{Li~\cite{li2017generative}} }&\textbf{ \textbf{Iizuka~\cite{iizuka2017globally}} }  &\textbf{\textbf{$L_1$+ GAN + ID} } &\textbf{Ours}  \\
\hline
\textbf{PSNR} &18.87 & 21.69 & 23.14  & 23.81 & 24.9\\
\textbf{SSIM} & 0.79 & 0.78 & 0.82  & 0.83 & 0.87\\
\hline
\end{tabular}
}
\vspace{-0.1in}
\end{table}

\vspace{-0.1in}

\section{Comparisons}
\vspace{-0.1in}
\label{sec:compare}
\paragraph{Comparisons with other methods}
We compare our result with other state-of-the-art inpainting frameworks.
%In particular, we crop and downsample our test image to $64\times 64$ when comparing with networks in~\cite{yeh2017semantic}. We use smaller mask when test on this model which is the only way we can conduct the two experiments. 
As shown in Figure~\ref{fig:comparison}, Pathak \etal \cite{pathak2016context} smoothly fill the missing part without any semantic meaning. 
Though Yeh et al. ~\cite{yeh2017semantic} can produce content with semantic meanings, their results tend to be blurry and fail to be spatially coherent with surrounding context.
%Yeh Pathak \etal in ~\cite{yeh2017semantic} fill the region with semantic meaning. 
Li \etal~\cite{li2017generative} and Iizuka \etal~\cite{iizuka2017globally} demonstrate sharper results but the results are still blurry and the generated facial features tends to be distorted and appear unnaturally with unmasked regions. 
Comparing to other methods, our approach is capable of generating high-fidelity facial
details with coherent blending with backgrounds. In addition, we successfully preserve the identity, providing result close to the ground truth.

\begin{figure*}[t]
	\centering
   \includegraphics[width=0.8\linewidth]{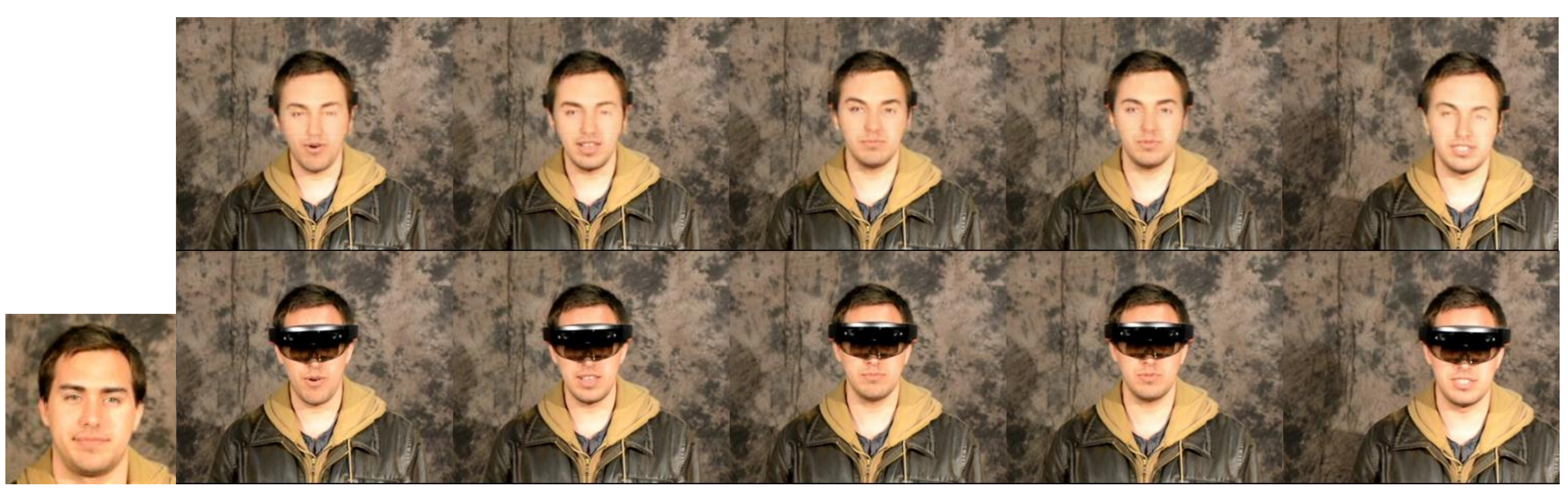}
   \vspace{0.07in}
   \caption{Reconstruction of real video data with frontal head pose. The first and second rolls are the selected frames reconstructed by our network and the ground truth. The leftmost image is used as reference for all the frames.}
\vspace{-0.1in}
\label{fig:Pose_uk}
\end{figure*}

\begin{figure*}[t]
	\centering
   \includegraphics[width=0.58\linewidth]{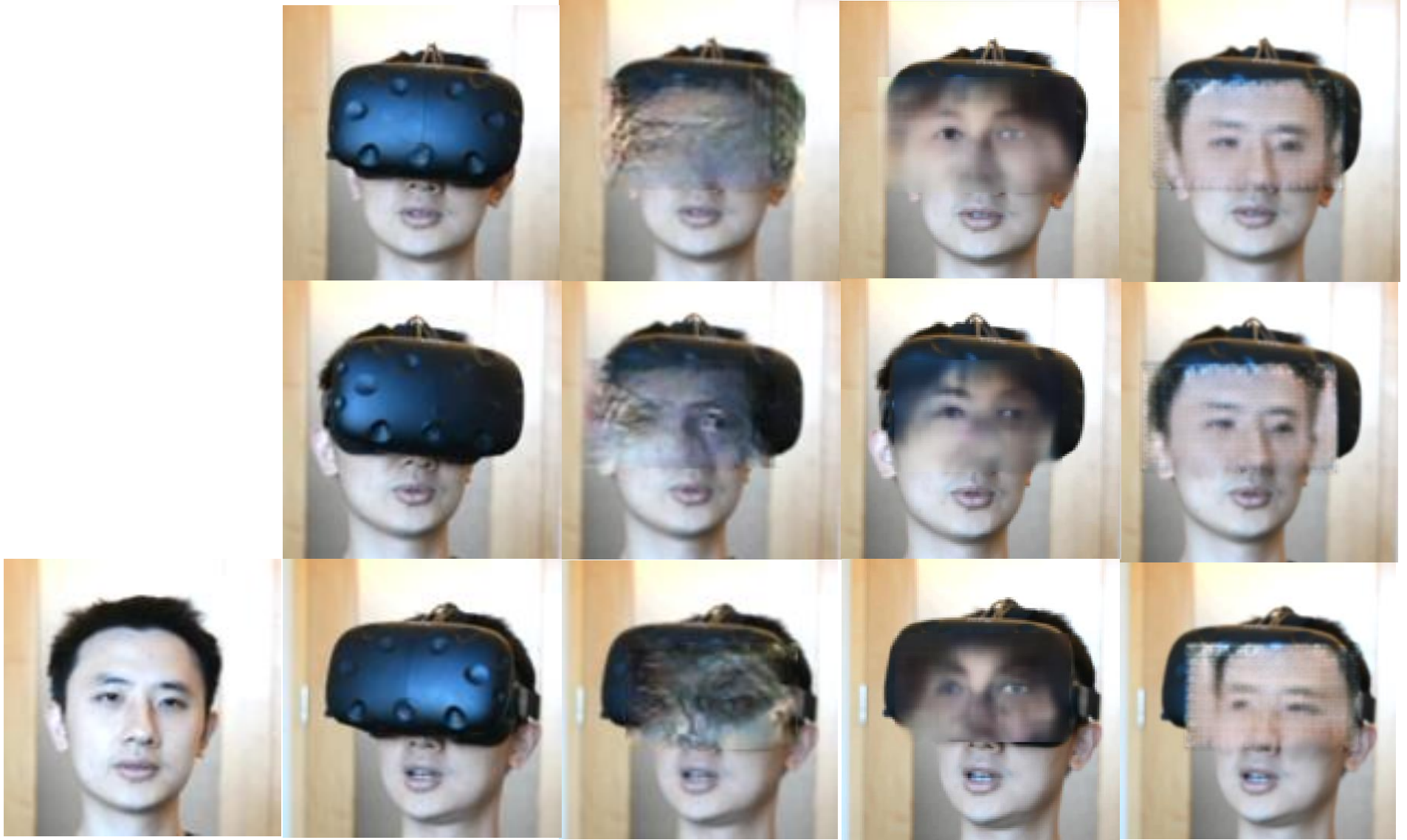}
   \vspace{0.07in}
   \caption{Reconstruction of real video data with large head pose variation when wearing VR/AR headsets. 
From the left to the right columns are: \textit{One single reference image}, \textit{Inputs}, \textit{Li \etal ~\cite{li2017generative}}, \textit{Iizuka \etal ~\cite{iizuka2017globally}} and \textit{Ours}.  The single reference image is used for all frames.}
\vspace{-0.1in}
\label{fig:Pose_wk}
\end{figure*}

\vspace{-0.2in}
\paragraph{Test on images with real occlusion.}
We test our network on video sequences in which the subjects are wearing HMDs. As we assume known head poses for our network. We need first to extract the head poses from occluded images. However, many HMDs, like HTC vive, provide real-time tracking of the head pose, which could be converted to our pose input via a simple calibration step.
We also train a pose prediction network using the synthetic data with known pose information. 
The network consists of 5 convolutional layers to extract the high-level features from the input image, and two fully connected layers to regress the feature into pose. In particular, our convolution part is same as the content prediction network of~\cite{YangLLSWL16}, which is trained to inpaint the missing content.
In Figure~\ref{fig:Pose_uk}, we show a video result with nearly frontal view, where we assume that pose is fixed to $(0,0,0)$. As seen from the results, our network produces stable results when the pose changes slightly.
In Figure~\ref{fig:Pose_wk}, we test our network on a video in which the subject is wearing a HTC vive VR headset and talking with large variations of head poses. 
%We used HTC vive and converted the provided head pose information to our pose input.
Despite the large head movement, our network can still generate very promising results.
Although our results of real data contain artifacts, the improvement is significant compared to Li \etal~\cite{li2017generative} and Iizuka \etal~\cite{iizuka2017globally}.
% Since the HMD causes a shadow area around the lips, our results also contain certain artifacts.

%-------------------------------------------------------------------------
\vspace{-0.1in}
\section{Conclusion, Limitation and Future work }
\label{sec:limit}

We present a novel learning-based approach for face inpainting with favorable property of preserving the identity of a given reference image. Furthermore, our approach offers
flexible pose control on the reconstruction results, making it possible to faithfully restore facial details in occluded video sequences with large face pose variations.
These two properties provide insight into solving the headset removal problem, which attracts increasing attention due to the surge of VR/AR techniques. 
%The key to our method is a generative neural network with dedicated loss constraints. In particular, we introduce the use of a reference identity image and pose control map. In generator, a reference network regularizes the identity loss between generated result and referenced identity. At discriminating side, a novel pose discriminator supervises the learning of correct pose transformation.
%In this paper, driven by application of HMD removal, we first observe the demanding of an identity preserved and pose robust face inpainting approach. Then we come up with a face inpainting network with a two novel components to handle the above issues. One component is the usage of reference image to provide the identity information. Another component is the incorporation of head pose, which allows the network to tolerate pose variations. Results show that our proposed network generate the best inpainting results with identity and pose coherence. 
Our network, in current form, cannot handle well extreme viewing angles and expressions  ({failure cases can be found in supplementary materials}). 
However, we believe that by including such cases in training dataset, the robustness of our network can be further improved.
In the video inpainting results, jittering can be observed in the transition between different frames as temporal coherency is not explicitly constrained in our formulation. 
Its worth investigating in the future work to add such additional constraint in our current framework.
The results of real data generated by our network still have some artifacts around the mask boundary. This is due to the fact that the lower face usually has shadows cast by the HMDs. One possible solution is to synthesize shadows for the training data.
It would also be an interesting future work to incorporate a pose estimation network to enable an end-to-end face inpainting network for videos.

\section{Acknowledgements}
We would like to show our gratitude to Nathan Jacobs, Xinyu Huang for sharing their pearls of wisdom with us during the course of this research, and we thank Qingguo Xu for assistance with experiments and comments that greatly improved the manuscript.

\bibliography{egbib,paper.bib}

\end{document}